\documentclass[runningheads]{llncs}
\usepackage[T1]{fontenc}
\usepackage{graphicx}
\usepackage{wrapfig}
\usepackage{multicol}
\usepackage{multirow}

\begin{document}
\title{Wide \& Deep Learning for Judging Student Performance in Online One-on-one Math Classes}
\titlerunning{Wide \& Deep Learning for Judging Student Performance}
%

\author{Jiahao Chen\inst{1}\orcidID{0000-0001-6095-1041} \and
Zitao Liu\inst{1}\orcidID{0000-0003-0491-307X}\thanks{The corresponding author: Zitao Liu} \and
Weiqi Luo\inst{2}\orcidID{0000-0001-5605-7397}}
\authorrunning{J. Chen et al.}
%
\institute{TAL Education Group, Beijing, China\\
\email{\{chenjiahao, liuzitao\}@tal.com}\and
Guangdong Institute of Smart Education, Jinan University, Guangzhou, China\\
\email{lwq@jnu.edu}}
\maketitle              
\begin{abstract}
In this paper, we investigate the opportunities of automating the judgment process in online one-on-one math classes. We build a Wide \& Deep framework to learn fine-grained predictive representations from a limited amount of noisy classroom conversation data that perform better student judgments. We conducted experiments on the task of predicting students' levels of mastery of example questions and the results demonstrate the superiority and availability of our model in terms of various evaluation metrics.

\end{abstract}
\vspace{-0.2cm}
\section{Introduction}
\vspace{-0.1cm}
\label{sec:intro}

Accurate student performance judgments provide essential information for teachers deciding what and how to teach and form the fundamental prerequisites for adaptive instructions \cite{sturmer2013declarative,dhamecha2018balancing,hao2021multi,liu2020personalized}. When inaccurate diagnoses and judgments are made of students' abilities, prior knowledge, learning and achievement motivation, or other student characteristics, teaching becomes less effective in terms of learning and achievement gains \cite{praetorius2017identifying,huang2020neural}. Our work is different from many existing studies along this direction \cite{praetorius2015judging,dhamecha2018balancing,haataja2019teacher,mcintyre2019capturing}. First, previous approaches try to manually conduct user studies to verify the accuracy of teacher judgment and the corresponding effects and influence. Second, the majority of existing works use the results of assignments or exams to represent student performance, which is a very lagged and coarse indicator.

In this paper, we aim to explore the opportunities of building an AI-driven model to perform automatic fine-grained student performance assessment by analyzing the in-class conversational data at the question level and aim to predict whether the student understands each example question on slides in the online one-on-one class. Our work focuses on utilizing the deep learning and neural network techniques to learn effective representations from a limited amount of noisy classroom conversation data that perform better student judgments. Our framework builds upon the Wide \& Deep learning architecture \cite{cheng2016wide} to achieve both memorization and generalization in one model by training a multi-layer perceptron (MLP) on representations from both wide and deep components. The wide component effectively memorizes sparse feature interactions between teachers and students, while deep neural networks can generalize and handle the linguistic variations in classroom conversations. Experiments show that the Wide \& Deep framework performs better than alternative methods when predicting student performance on a real-world educational dataset.

\vspace{-0.2cm}
\section{The Wide \& Deep Framework}
\vspace{-0.1cm}
\label{sec:method}

In this work, we design the Wide \& Deep learning framework to achieve both memorization and generalization in one model, by training a MLP model with representations from both the wide component and the deep component. 

\noindent \textbf{The Wide Component}. In the wide component, we handcraft 25 features to capture the teacher-student interactions on each example question in the online one-on-one math class. Our manually-engineered features include both continuous-valued features, i.e., conversation duration and Jaccard similarity between teacher's and student's sentences, and discrete-valued features, i.e., the number of words per conversation from the teacher, the number of student-spoken sentences with less than 3 words. We use one-hot encoding for discrete features and project them to dense embedding space with a linear projection matrix $\mathbf{W}$, i.e., $\mathbf{x}_d^* = \mathbf{W} \mathbf{x}_d$, where $\mathbf{x}_d$ represents the one-hot representations of discrete features and $\mathbf{x}_d^*$ represents the projected dense representations for discrete features. The overall features from the wide component $\mathbf{x}_{wide}$ is the concatenation of the original continuous features $\mathbf{x}_c$ and projected dense features $\mathbf{x}_d^*$, i.e., $\mathbf{x}_{wide} = \mathbf{x}_c \oplus \mathbf{x}_d^*$ where $\oplus$ denotes the concatenation operation. A two-layer fully connected deep neural network is trained on features from wide component $\mathbf{x}_{wide}$.

\noindent \textbf{The Deep Component}. We design a deep component for learning effective low-dimensional representations of the rich in-class conversational information and teaching styles and instructions variations. Firstly, we use the ``EduRoBERTa'' \cite{EduRoBERTa} to obtain the sentence embedding $\mathbf{e}_i$ of each spoken transcription sentence $\mathbf{s}_i$, i.e., $\mathbf{e}_i = \mbox{EduRoBERTa}(\mathbf{s}_i)$. Then, we use a deep sequential model to capture the relation of long-range dependencies among the sentences in the original conversational texts. In this work, we use bidirectional LSTM as our sentence encoder, i.e., $\{\mathbf{h}_1, \mathbf{h}_2, \cdots, \mathbf{h}_{n}\} = \mbox{BiLSTM}(\{\mathbf{e}_1, \mathbf{e}_2, \cdots, \mathbf{e}_{n}\})$. Furthermore, we apply the structured self-attention mechanism to each learned representation from BiLSTM, i.e., $\{\mathbf{q}_1, \mathbf{q}_2, \cdots, \mathbf{q}_{n}\} = \mbox{Attention}(\{\mathbf{h}_1, \mathbf{h}_2, \cdots, \mathbf{h}_{n}\})$. The structured self-attention is an attention mechanism relating different positions of a single sequence in order to compute a representation of the sequence \cite{vaswani2017attention}. At the end, we use the average pooling operation to get the low-dimensional dense representation $\mathbf{x}_{deep}$ from the deep component, i.e., $\mathbf{x}_{deep} = \mbox{MeanPooling}(\{\mathbf{q}_1, \mathbf{q}_2, \cdots, \mathbf{q}_{n}\})$. A four-layer fully connected deep neural network is trained on features from wide component $\mathbf{x}_{wide}$.

\noindent \textbf{Joint Learning}. The wide component and deep component are combined using an element-wise sum of their output log odds as the prediction, which is then fed to a Softmax function to output the probability score of the student's level of mastery of each example question. Our Wide \& Deep Model is learned by minimizing the standard cross-entropy loss function.

\vspace{-0.2cm}
\section{Experiments}
\vspace{-0.1cm}
\label{sec:experiment}

In this work, we collect 5226 unique samples of teacher-student conversations of solving example question from 497 recordings of online one-on-one math classes in grade 8 from a third-party K-12 online education platform. We use a third-party publicly available ASR service for the conversation transcriptions. We evaluate the performance of our proposed approach (denoted as \textbf{W\&D}) by comparing it with the following baselines: (1) \textbf{GBDT}: Gradient boosted decision tree \cite{friedman2002stochastic} on handcrafted wide features; (2) \textbf{LSTM w. EduRoBERTa}: Similar to the proposed approach but it eliminates the contribution of features from the wide component. We utilize the XGBoost library \cite{chen2016xgboost} for the GBDT training where we grid search the max\_depth over $[1, 2, \cdots, 9]$ and n\_estimators over $[10, 20, \cdots, 90]$ and with a fixed subsample of 0.9. For the BiLSTM model, the size of hidden states is grid searched over $[64,128,256]$. We set batch size to 256 and conduct the hyper-parameter tuning by selecting the one with the highest accuracy on the validation set and we report the model performance on the test set.

\begin{table}[!hbpt]
\centering
\vspace{-0.6cm}
\caption{Overall performance in terms of accuracy, micro-F1 and macro-F1.}\vspace{-0.1cm}
\label{tab:overall}
\begin{tabular}{l|c|c|c}
\hline
Model Name            & Accuracy   & micro-F1 & macro-F1 \\
\hline
GBDT                   & 0.705          & 0.676          & 0.606          \\
LSTM w. $\mbox{EduRoBERTa}$ & \textbf{0.728}          & 0.704          & 0.643          \\
W\&D(Ours)               & \textbf{0.728}          & \textbf{0.709}          & \textbf{0.649}          \\

\hline
\end{tabular}
\vspace{-0.6cm}
\end{table}

We list the test results in terms of accuracy and F1 scores from both micro and macro perspectives \cite{van2013macro}, shown in Table \ref{tab:overall}. We make the following two observations: (1) The LSTM w. $\mbox{EduRoBERTa}$ led to a macro F1 of 0.643, which is a notable improvement over the performance of GBDT. It showed that text information is important for judging student performance; and (2) our W\&D model achieves the best performance in all metrics by combining wide and deep features. This suggests that although the deep model has pretty good than GBDT, the wide features can improve the performance of LSTM. Thus, it is essential to incorporate both wide and deep features.

\vspace{-0.2cm}
\section{Conclusion}
\vspace{-0.1cm}
\label{sec:conclusion}
In this paper, we present a Wide \& Deep framework to predict students' level of mastery of questions. Combining the handcraft features with a deep learning-based model can improve the performance. As part of future work, we plan to find more efficient method to encode the conservation text of students and teachers. Furthermore, we will do a more profound analysis of handcraft features, which can help us understand which features are essential for judging students' performance.

\vspace{-0.3cm}
\subsubsection{Acknowledgements} This work was supported in part by National Key R\&D Program of China, under Grant No. 2020AAA0104500; in part by Beijing Nova Program (Z201100006820068) from Beijing Municipal Science \& Technology Commission and in part by NFSC under Grant No. 61877029.

\vspace{-0.2cm}
%
%

\bibliographystyle{splncs04}
\bibliography{aied2022}

\end{document}